\documentclass{article}

    \PassOptionsToPackage{numbers, compress}{natbib}

    \usepackage[preprint]{neurips_2020}

\usepackage{floatrow}
\usepackage[utf8]{inputenc} 
\usepackage[T1]{fontenc}    
\usepackage{hyperref}       
\usepackage{url}            
\usepackage{booktabs}       
\usepackage{amsfonts}       
\usepackage{nicefrac}       
\usepackage{microtype}      
\usepackage{amsmath}
\usepackage{natbib}
\usepackage{algorithm}
\usepackage{booktabs} 
\usepackage{wrapfig}
\usepackage{adjustbox}
\usepackage{subfigure}
\usepackage{float}
\usepackage{sidecap}

\def\eg{\emph{e.g.~}}

\def\ie{\emph{i.e.~}}

\usepackage{mathtools}

\DeclareMathOperator*{\argmax}{arg\,max}
\DeclarePairedDelimiter\floor{\lfloor}{\rfloor}

\title{Enhancing sensor resolution improves CNN accuracy given the same number of parameters or FLOPS}

\author{
Ali Borji \\
\texttt{aliborji@gmail.com} 
\thanks{Code is available at: \url{https://github.com/aliborji/resolution.git}}
}

\begin{document}

\maketitle

\begin{abstract}
High image resolution is critical to obtain a good performance in many computer vision applications\footnote{Human visual apparatus is very sophisticated. Our eyes capture very high resolution images, perhaps much higher than the resolution of best cameras today. The resolution at the fovea is higher than the visual periphery. High image resolution might be one of the key reasons why human vision is so good.}. Computational complexity of CNNs, however, grows significantly with the increase in input image size. Here, we show that it is almost always possible to modify a network such that it achieves higher accuracy at a higher input resolution while having the same number of parameters or/and FLOPS. The idea is similar to the EfficientNet paper but instead of optimizing network width, depth and resolution simultaneously, here we focus only on input resolution. This makes the search space much smaller which is more suitable for low computational budget regimes. More importantly, by controlling for the number of model parameters (and hence model capacity), we show that the additional benefit in accuracy is indeed due to the higher input resolution. Preliminary empirical investigation over MNIST, Fashion MNIST, and CIFAR10 datasets demonstrates the efficiency of the proposed approach.
\end{abstract}

\section{Introduction}

Modern deep neural networks are very expensive to train and only a few big corporations and universities can afford to scale them up. To maintain a reasonable cost, CNNs have been traditionally applied to images with resolution of about 200-300 pixels. Increasing the input resolution increases the computational complexity dramatically. For example, a ResNet-152 network~\cite{he2016deep} with input size $224 \times 224$ has 11,558,837,248 MACs and 60,192,808 parameters. Doubling the input size for this network quadruples the number of MACs while keeping the number of parameters the same.

In~\cite{tan2019efficientnet}, Tan and Le proposed a method, called EfficientNet, to systematically scale up CNNs. They showed that carefully balancing network depth, width, and resolution can lead to better performance in object recognition and object detection (EfficientDet) tasks. They perform a grid search in these dimensions, train a model for each combination, and find the one that leads to the best performance. Since this is a very expensive process they only performed the gird search at the first stage (using a baseline architecture) and used the same solution to build the subsequent models (\ie EfficientNet-B0 to EfficientNet-B7). The networks in the EfficientNet paper are constrained to be within a certain budget (\eg two times more FLOPS than the original network FLOPS) and all components of the base network can be altered, which makes the search space very large. 

The problem of interest here is similar to the EfficientNet setup with two distinctions. First, {\bf we seek to answer whether it is the higher model capacity (as a result of higher input resolution) or the higher resolution itself that is responsible for performance improvement}. 
EfficientNet paper does not answer this question since increasing image resolution leads to a higher number of parameters, and hence higher network capacity. Therefore, the improved accuracy can be merely due to the higher number of model parameters rather than the pure effect of image resolution. Here, we propose methods to keep the number of parameters or FLOPS the same. The search space in our solutions is smaller than the search space in EfficientNet which makes our approach more suitable for practical purposes. Second, our approaches allow improving the accuracy with the fixed number of FLOPS. This is desirable for scenarios in which additional computational budget is not an option and therefore a model should be optimized to best utilize the existing resources.

\begin{figure}[t]
{ \caption{{A pilot experiment over MNIST, F-MNIST, and CIFAR datasets investigating the impact of image resolution on the number of parameters and FLOPS. We used a CNN with two convolution, two pooling, and two fc layers. Over CIFAR-10, we varied image size ($\in [8,16,32]$), conv filter size ($\in [3,5,7]$) as well as pooling filter size ($\in [1,2]$). Over MNIST and F-MNIST, we varied image size ($\in [7,14,28]$), size of conv filters ($\in [2,3,5]$) as well as pooling filter sizes ($\in [1,2]$). Notice that some combinations were discarded since they did not lead to a valid configuration. Each dot in the plots indicates a single model. Networks for CIFAR and MNIST are the same except that the former uses 30 conv filters in each of conv layers whereas the latter uses 10.}}
\label{fig:illust} }
{
\includegraphics[width=.7\textwidth]{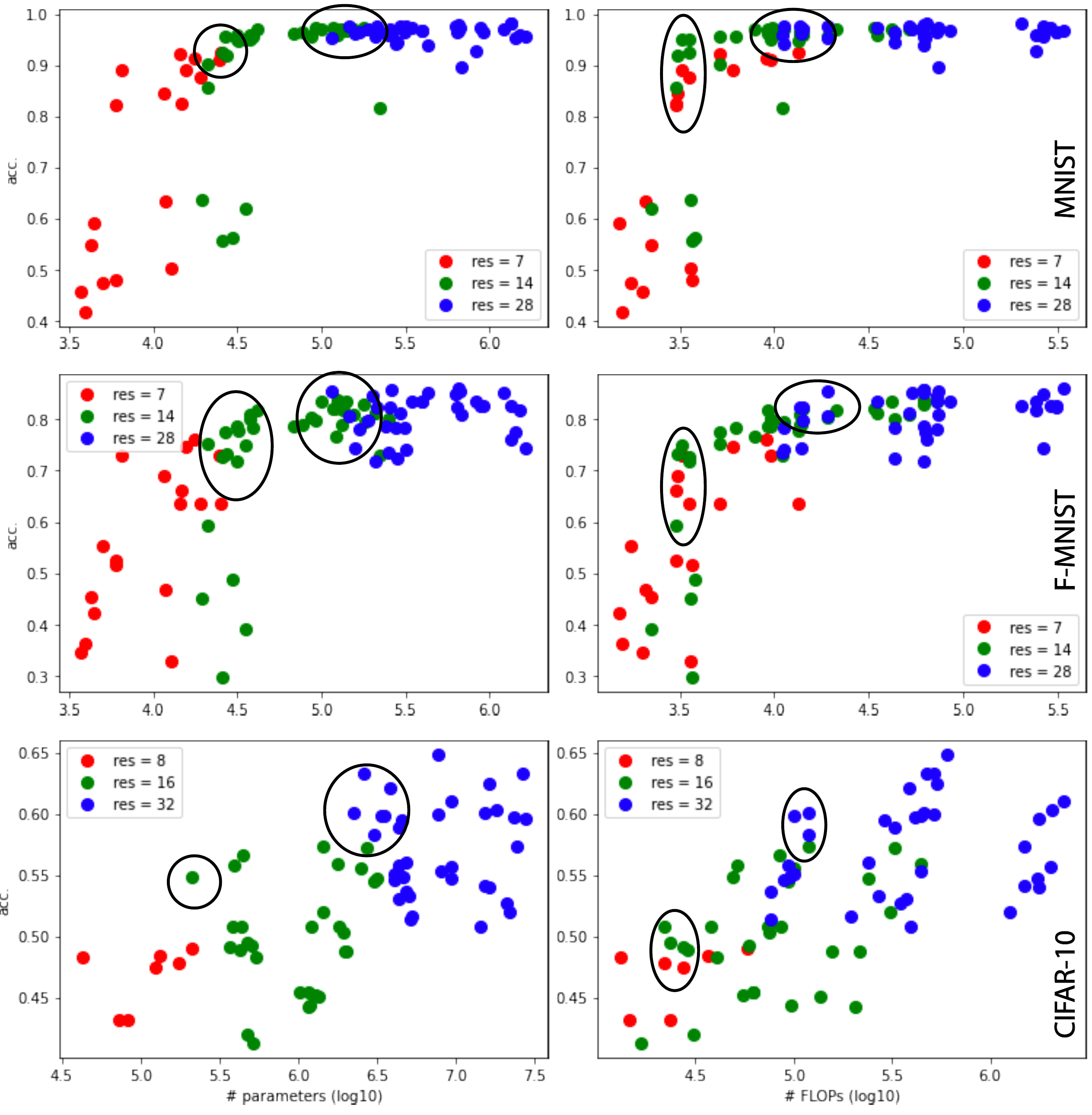}}
\end{figure}

To illustrate the idea, we trained several LeNet CNNs (conv1 $\Rightarrow$ pool1 $\Rightarrow$ conv2 $\Rightarrow$ pool2 $\Rightarrow$ fc1 $\Rightarrow$ fc2) with various input resolutions. In Fig. \ref{fig:illust}, model accuracy is plotted against the number of parameters (left column) and FLOPS (right column). It can be seen that a) increasing input resolution improves the accuracy (as expected!), and b) looking vertically at certain positions in the $x$ axis (coincides with where the resolution changes), it is often possible to find a model that performs better at a higher resolution than a model at lower resolution (with about the same number of parameters or FLOPs). This observation suggests that for a model with resolution $R0$, we may be able to find a new model with resolution $R1>R0$ such that it performs better with the same number of FLOPS or parameters. We explore this possibility and propose methods to search for such models.

\noindent \textbf{Problem statement.} 
Given a network with a predefined architecture, we wish to find a network with higher input resolution that has the best accuracy with the same number of parameters or FLOPS as the original network\footnote{The constraint can also be formulated as $0 < D(M_i) - D(M) < \alpha$ where $\alpha$ is the additional budget.}. Formally, 
\begin{equation}
 M^* = \argmax_i \ P(M_i) \ \  \text{s.t} \ \ D(M_i) = D(M) 
 \label{eq:formal}
\end{equation}
where $D(.)$ is the desired property (\eg number of parameters or FLOPS), and $P(.)$ is model accuracy. $M_i$ denotes the model with a higher input resolution\footnote{One may choose to vary other dimensions such as network width or depth.} than the original model. We will have a solution when $P(M^{*}) > P(M)$. 
Since the search space\footnote{\ie the space of all model architectures that satisfy the condition in Eq.~\ref{eq:formal}.} is usually very large, here we propose four approaches that can be utilized to modify an arbitrary network to accept a higher input resolution. In practice, one can also search in the space of resolutions and look for the one that improves the performance (as we will do in the Experiments section).

\section{Background}

\noindent {\bf Number of layer parameters:}
Consider a conv layer with $K$ filters each of size $C \times C \times D$ (assume no params for nonlinearity), then the number of parameters to learn are:
\begin{equation}
    Params = K C^2 D + K
\end{equation}
With no bias, it reduces to $K C^2 D$.

A pooling layer (max or avg) has no parameters. A fully connected layer with input layer size $I$ and output layer size $O$ has the following number of parameters:
\begin{equation}
    Params = IO + O
\end{equation}
With no bias, it reduces to IO.

\noindent {\bf Size of the output map for conv and pooling layers:}
Output size of a convolution or pooling layer with input size $N$, filter size
$C$, padding $P$, stride $S$, and dilation $D$ (default =1) is:

\begin{equation}
    M = \floor*{\frac{N - D (C - 1) - 1 +  2P}{S} + 1}
\end{equation}

\noindent {\bf Number of FLOPS\footnote{Our calculation is only based on inference computations and gradient computations during backpropagation are discarded. Some libraries such as \url{https://github.com/Lyken17/pytorch-OpCounter} compute MAC(multiply-accumulate) which is lower than the number of FLOPS since one FLOP contains one multiplication and one addition. Sometimes one application of kernel plus bias is considered one MAC (\ie dot product plus bias addition).}} Consider a conv layer with output map size $M$, squared kernel with size $C$, $K_{in}$ number of input channels, and $K_{out}$ number of output channels. Assume that convolution is implemented as sliding window and non-linearity is done for free. The number of FLOPS is:
\begin{eqnarray}
    FLOPS = M^2(C^2K_{in} + 1)K_{out},
    \label{eq:FLOPS}
\end{eqnarray}
corresponding to the number of multiplications, additions and bias addition terms. Assuming no bias, we have $M^2C^2K_{in}K_{out}$ FLOPS. Above equation includes every pixel in the output map and hence implicitly takes input map size, stride and padding into account (\ie no need to mention them explicitly).

Pooling layers are usually discarded in FLOPS computation since they have negligible number of FLOPS. For example, a max pooling layer with filter size 2 and stride 2 on a 112$\times$112$\times$128 input tensor (WHD) has 112$\times$112$\times$128 = 1,605,632 FLOPS or around 1.6 mega FLOPS which is very small compared to the 100s of mega FLOPS in convolution or fully-connected layers.

A fc layer with input size $I$ and output layer size $O$ has $FLOPS = IO + O$ (same as the number of parameters). Assuming no bias, it is lowered to $IO$.

\begin{figure}[ht]
  \includegraphics[width=.35\linewidth]{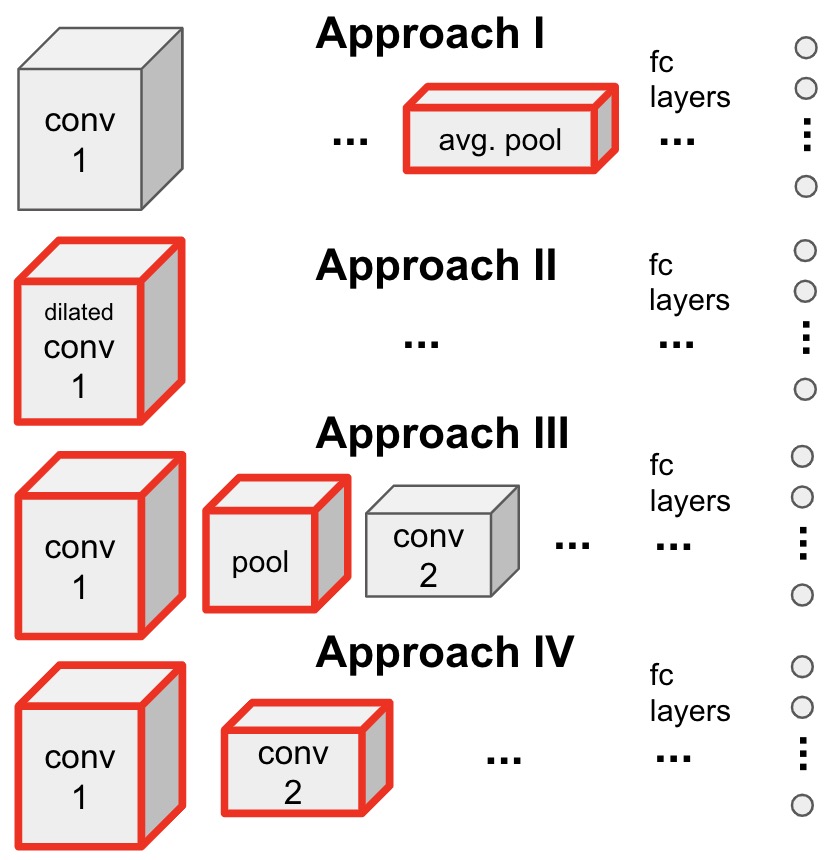}
    \caption{Four approaches to compensate for the resolution enhancement. Red border cubes indicate modified or added components. }
  \label{fig:approaches}
\end{figure}

\section{Approaches}
Assume the input shape is $N \times N \times D$. We aim to increase the resolution to $N^{'} \times N^{'} \times D$ ($N^{'} > N)$ such that the total number of model parameters and/or FLOPS stays the same. There are at least four easy ways to modify the network structure to fulfill this goal (See Fig.~\ref{fig:approaches}), as is described below. 

{\bf An important point to notice here is the need to increase the filter size with a bigger input size. This is because the old kernel size may be too small to capture useful structural information.}

\subsection{Approach I: Average pooling layer before the fc layers}
The idea is to use an average pooling layer right before the first fc layer. This way any input resolution can be handled. Regardless of the input size, the average pooling reduces each map to a single number and hence the input tensor to the first fully connected will always have the same shape (a 1D vector). Despite being simple, however, this approach has the drawback of losing a lot of spatial information. 

Using average pooling keeps the number of parameters fixed but increases the number of FLOPS. Since the increase in the number of FLOPS is negligible, in practice it can be discarded. If need be, however, it is possible to adjust the network to keep the number of FLOPS the same. One must solve for:
\begin{equation}
\text{Find} \ \  Z_1^{'} \ \ \  \text{s.t} \ \  \ U_1^2 V_1 + V_1 Z_1 + Z_1 Z_2 =  U_1^{'2} V_1 + V_1 Z_1{'} + Z_1{'}Z_2
\end{equation}
where $U_1 \times U_1 \times V_1$ and $U_1^{'} \times U_1^{'} \times V_1$ are the shapes of the average pooling layer in the original and new networks, respectively. $Z_1$ and $Z_1^{'}$ are the size of the fc layers following the pooling layer in the original and new networks, respectively. $Z_2$ is the fc layer after $Z_1$. The idea is to vary the number of neurons in the $Z_1$ fc layer to compensate for the increase in FLOPS ($U_1^{'} > U_1,  Z_1^{'} < Z_1$). Notice that the final solution will have both the same number of parameters and FLOPS as the original network. We assumed no bias here.

\subsection{Approach II: Dilated convolution in the first conv layer}
Perhaps the simplest approach is to use dilated convolutions. Dilated convolution (a.k.a atrous convolution) controls the spacing between the kernel points\footnote{See https://towardsdatascience.com/a-comprehensive-introduction-to-different-types-of-convolutions-in-deep-learning-669281e58215}.
The idea here is to use dilated convolution to cover a wider area in the image (\ie increased receptive field) while keeping the number of parameters the same since dilation does not add new parameters (nor FLOPS). It does, however, increase the output map size which we have to fix. To do so, we have to solve for new stride, padding and dilation to keep the conv1 output resolution the same as before (and hence the rest of the network intact). Formally, we need:
\begin{eqnarray}
    M^{'}_1 = M_1 \Rightarrow \floor*{\frac{N^{'}_1 - D^{'}_1 (C_1 - 1) - 1 +  2P^{'}_1}{S^{'}_1} + 1} =  \floor*{\frac{N_1 - C_1 + 2P_1}{S_1} + 1},
    \label{eq:dilated}
\end{eqnarray}
where $M^{'}_1$ is the new conv1 output size and $N^{'}_1$ is the new resolution ($N^{'}_1$ is given). We assume that the original network does not have dilation (\ie dilation = 1). Notice that introducing dilation does not change the kernel size. Eventually, the solution is:
\begin{eqnarray}
    \text{Find} \ \  P_1^{'}, S_1^{'}, D^{'}_1 \ \ (\in \mathbb{N}) \nonumber \\ 
    \text{s.t} \ \ M^{'}_1 = M_1  
\end{eqnarray}

{\bf The first two approaches do not allow the filter size to grow. This might be detrimental when dealing with higher resolution images where structures appear larger than before. The following two approaches remedy this shortcoming by increasing the filter size.}

\subsection{Approach III: Pooling layer after the first conv layer}
The idea here is to increase the filter size in the first conv layer but reduce the number of its filters to compensate for the added parameters (or FLOPS). This causes the resolution mismatch for the second layer which we have to fix. To remedy, we append a new pooling layer (recall that pooling does not introduce parameters). 

\noindent  {\bf Keeping the number of parameters the same.} 
Assume the first conv layer has $K_1$ filters, each with size $C_1 \times C_1 \times D$ ($D=3$ for RGB images). Thus, the number of parameters are $K_1 C_1^2D $ (no bias). The number of parameters in the new network are $K_1^{'} C_1^{'2}D $ where $'$ indicates new architecture. Since we desire the number of parameters to be the same, we must have:
\begin{equation}
    K_1 C_1^2 D  = K_1^{'}C_1^{'2}D  \ \ \text{where} \ \ C_1^{'} > C_1 \\
    \label{eq:conv1_params}
\end{equation}
Thus: 
\begin{equation}
\frac{K_1^{'}}{K_1} = \big(\frac{C_1}{C_1^{'}}\big)^2
\end{equation}
which means to increase the filter size we have to quadratically lower the number of filters, \ie $K_1^{'} < K_1$. To fix the resolution (\ie before feeding to the subsequent layers), we should find stride $S_p$, padding $P_p$, and pooling kernel size $C_p$ such that the new resolution $M_p $ is equal to $M_1$:
\begin{eqnarray}
    M_1^{'} = \floor*{\frac{N^{'} - C_1^{'} + 2P_1^{'}}{S_1^{'}} + 1} \\
    M_p = \floor*{\frac{M_1^{'} - C_p + 2P_p}{S_p} + 1} 
\end{eqnarray}

where $M_1^{'}$ is the output size of modified conv1 layer. Ultimately, we should solve for:
\begin{eqnarray}
    \text{Find} \ \ K_1^{'}, C_1^{'}, P_1^{'}, S_1^{'}, C_p, P_p, S_p (\in \mathbb{N}) \nonumber \\ 
    \text{s.t} \ \ M_p = M_1, \  \text{Eq.} \  \ref{eq:conv1_params} \text{, and} \ C_1^{'} > C_1
    \label{eq:apprpach3_1}
\end{eqnarray}

\noindent {\bf Keeping the number of FLOPS the same.}
The solution is as in Eq.~\ref{eq:apprpach3_1} but with a different condition:
\begin{eqnarray}
    \text{Find} \ \ K_1^{'}, C_1^{'}, P_1^{'}, S_1^{'}, C_p, P_p, S_p (\in \mathbb{N}) \nonumber \\ 
    \text{s.t} \ \ M_p = M_1, \ M_1^2C_1^2DK_1 = M_1^{'2}C_1^{'2}DK_1^{'} +  M_1^{'2}K_1^{'} \text{, and} \ C_1^{'} > C_1 
    \label{eq:apprpach3_2}
\end{eqnarray}
In this new condition, in addition to resolution the number of FLOPS should stay the same. $M_1^{'2}K_1^{'} $ is the additional FLOPS introduced by the pooling layer.

\subsection{Approach IV: Adjusting the first two conv layers}
Here, we adjust the parameters of the first two conv layers (no added layer) to construct new models.

\noindent  {\bf Keeping the number of parameters the same.} 
Increasing the filter size in the first conv layer increases the number of parameters. To compensate for this, we should change the second layer as well. 
The adjustment is similar to above. Assume the second conv layer has $K_2$ filters of shape $C_2 \times C_2 \times K_1$, where $K_1$ is the number of filters in the first conv layer. Thus, the number of parameters in the second layer are $K_1 K_2 C_2^2$. We need to have:
\begin{eqnarray}
\label{eq:approach4_1}
 K_1^{'}C_1^{'2}D + K_2 C_2^{'2}K_1^{'} = K_1 C_1^2D + K_2 C_2^{2}K_1, 
 \end{eqnarray}
which forces the total number of parameters in the first two layers remain fixed. Notice that we need to keep the number of filters in the second conv layer the same as before so that the rest of the network is not impacted. We also have to make sure the output resolution of conv2 is as before:
\begin{eqnarray}
\label{eq:approach4_2}
    M_2 = \floor*{ \frac{M_1 - C_2 + 2P_2}{S_2} + 1}  \\ 
    M_2^{'} = \floor*{ \frac{M_1^{'} - C_2^{'} + 2P_2^{'}}{S_2^{'}} + 1}
\end{eqnarray}
Finally, the solution is as follows:
\begin{eqnarray}
    \text{Find} \ \  K_1^{'}, C_1^{'}, P_1^{'}, S_1^{'}, C_2^{'}, P_2^{'}, S_2^{'} \ \ (\in \mathbb{N}) \\ \nonumber
    \text{s.t} \ \   \ M_2^{'} = M_2,  \  \text{Eq. \ref{eq:approach4_1},} \  \text{and} \ C_1^{'} > C_1 \ 
\end{eqnarray}

\noindent {\bf Keeping the number of FLOPS the same.}
The objective here is to keep the total number of FLOPS the same in the first two layers:
\begin{equation}
    D M_1^{'2}C_1^{'2}K_1^{'} + M_2^{'2}K_1^{'}C_2^{'2}K_2 = D M_1^2C_1^2K_1 +  
     M_2^2K_1C_2^2K_2 
    \label{eq:flops}
\end{equation}

As above, we have to also fix the resolution after the second layer as well. Finally, we have:
\begin{eqnarray}
    \text{Find} \ \  K_1^{'}, C_1^{'}, P_1^{'}, S_1^{'}, K_2^{'}, C_2^{'}, P_2^{'}, S_2^{'} \ \ (\in \mathbb{N}) \nonumber \\  
    \text{s.t}  \ \  \  M_2^{'} = M_2, \  \text{Eq. \ref{eq:flops}} \text{, and} \ C_1^{'} > C_1 
\end{eqnarray}

\section{Results}

The approaches presented above find solutions for a desired higher resolution. In practice, one can vary the resolution and see which one works better (\ie also including the resolution dimension as part of the grid search). Here, we consider 
$N < N^{'} \le 2N$. We vary the kernel size in the range 3:8, stride in 1:5, padding in 1:5, and dilation in 2:4. To conduct the experiments, we first down sample the images and then increase the resolution up to the maximum resolution. 

Network architectures are the same as the ones used to produce the results in Fig.~\ref{fig:illust}, except that we drop the pooling layers after conv layers for conducting approach IV. For each resolution, we first sample parameters to construct a number of original networks (=10). For each original network, we then find a list of solutions (with higher resolution input) and randomly choose four networks from this list. Recall that the original network and its corresponding solutions all have the same number of parameters\footnote{We are considering equal parameter case here.}. Each network (original or solution) are then trained and tested three times, and the average test accuracy is calculated. The maximum performance among the solution is recorded (See Fig.~\ref{fig:Results2}). 

\begin{figure}[t]

\includegraphics[width=\linewidth]{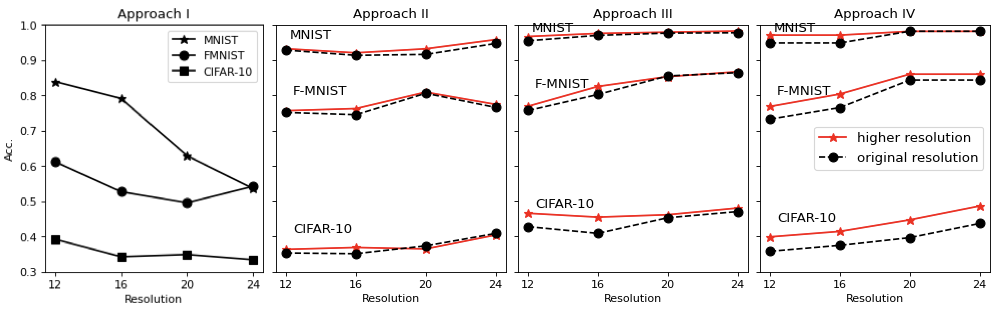}
    \caption{Results of the proposed approaches. The black and red dots represent the performance of the original model (averaged over 10 different models; Fig.~\ref{fig:Results2}) and the performance of the best solution with the same number of parameters corresponding to each original model (\ie average of the max performances). In approach I, the performance drops as the resolution grows. This is because average pooling collapses a map to a single number thus losing a lot of information. Notice that performances can not be compared across approaches since models are sampled randomly.}
  \label{fig:Results}
\end{figure}

\begin{figure}[t]

\includegraphics[width=\linewidth]{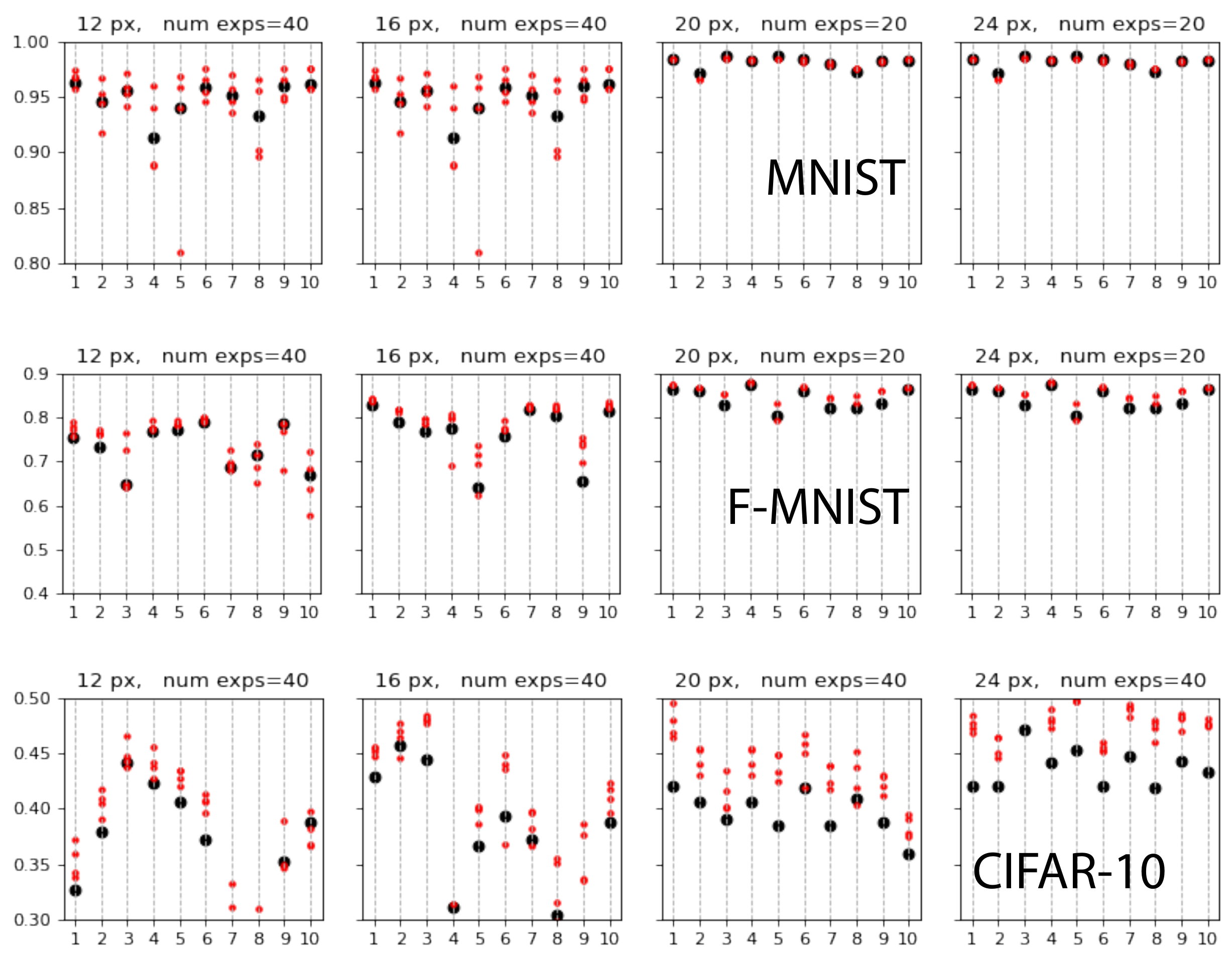}
    \caption{Accuracy of 10 individual models (randomly sampled architectures) corresponding to the right-most panel in Fig.~\ref{fig:Results} (\ie Approach IV). The large black dot represents the original model at the resolution specified on each panel caption. The red dots represent the matching models (same number of parameters as the original models). As you can see it is often possible to find a new model that works better at a higher resolution. Here, we only considered a small number (=4) of matching models with the original model. The results in Fig.~\ref{fig:Results} are averaged over 10 models at each resolution (\ie each panel is summarized). ``num exp" on the panel caption shows the total number of models that were matched across 10 original models. Notice that original models are different from each other and their parameters (filter channels, stride, etc) are also sampled.}
  \label{fig:Results2}
\end{figure}

Fig.~\ref{fig:Results} shows the results (averaged over 10 original models). Approach I is not effective since average pooling severely hinders the accuracy. Approaches II and III show a marginal improvement. Approach IV does the best and is more effective at lower resolutions (since there is more opportunity to increase the resolution). It performs well across all three datasets. {\bf We thus conclude that higher image resolution, independent of model capacity, improves the model accuracy.}

\section{Discussion}
The proposed approaches are in spirit similar to the EfficientNet work and are essentially a type of network architecture search (NAS). An important finding that is worth digging deeper is that among networks with the same model capacity (number of parameters), the ones with the higher input resolution usually perform better. This means that higher image resolution provides fine grained information that can be exploited by the network. An important lesson from a practical point of view is that given a limited computational budget (FLOPS), it might be possible to gain better accuracy by increasing the image resolution. 

Future work can apply these ideas to other types of networks such as CapsuleNets and Transformers, as well as other tasks such as object detection 
and semantic segmentation where high image resolution is critical. Here, we solved for the number of parameters and FLOPS one at a time. It is easy to extend the approaches to include both conditions (\ie joint optimization). Further, depending on the available computational resources, one can choose to adjust more number of layers.

\bibliographystyle{plain}
\bibliography{references}

\begin{thebibliography}{1}

\bibitem{he2016deep}
Kaiming He, Xiangyu Zhang, Shaoqing Ren, and Jian Sun.
\newblock Deep residual learning for image recognition.
\newblock In {\em Proceedings of the IEEE conference on computer vision and
  pattern recognition}, pages 770--778, 2016.

\bibitem{tan2019efficientnet}
Mingxing Tan and Quoc Le.
\newblock Efficientnet: Rethinking model scaling for convolutional neural
  networks.
\newblock In {\em International Conference on Machine Learning}, pages
  6105--6114. PMLR, 2019.

\end{thebibliography}
\end{document}